\documentclass[runningheads]{llncs}
\usepackage[T1]{fontenc}
\usepackage{graphicx}
\usepackage{multirow}
\usepackage{xcolor}
\usepackage{hyperref}

\begin{document}

\title{CXR-CLIP: Toward Large Scale Chest X-ray Language-Image Pre-training}

\author{Kihyun You\inst{1} \and
Jawook Gu\inst{1} \and
Jiyeon Ham\inst{1} \and
Beomhee Park\inst{1} \and
Jiho Kim\inst{1} \and
Eun K. Hong\inst{1} \and
Woonhyuk Baek\inst{1} \and
Byungseok Roh\inst{1}
}
% index{You, Kihyun}
% index{Gu, Jawook}
% index{Ham, Jiyeon}
% index{Park, Beomhee}
% index{Kim, Jiho}
% index{Hong, Eun}
% index{Baek, Woonhyuk}
% index{Roh, Byungseok}

\authorrunning{K. You et al.}
\institute{Kakaobrain, Seongnam, Republic of Korea  \\
\email{\{ukihyun, jawook.gu, jiyeon.ham, brook.park, tyler.md, amy.hong, wbaek, peter.roh\}@kakaobrain.com}\\}
\maketitle              % typeset the header of the contribution
\begin{abstract}
A large-scale image-text pair dataset has greatly contributed to the development of vision-language pre-training (VLP) models, which enable zero-shot or few-shot classification without costly annotation. However, in the medical domain, the scarcity of data remains a significant challenge for developing a powerful VLP model. In this paper, we tackle the lack of image-text data in chest X-ray by expanding image-label pair as image-text pair via general prompt and utilizing multiple images and multiple sections in a radiologic report. We also design two contrastive losses, named ICL and TCL, for learning study-level characteristics of medical images and reports, respectively. Our model outperforms the state-of-the-art models trained under the same conditions. Also, enlarged dataset improve the discriminative power of our pre-trained model for classification, while sacrificing marginal retrieval performance. Code is available at \url{https://github.com/kakaobrain/cxr-clip}.

\end{abstract}
\section{Introduction}
Chest X-ray (CXR) plays a vital role in screening and diagnosis of thoracic diseases~\cite{world2016communicating}.
The effectiveness of deep-learning based computer-aided diagnosis has been demonstrated in disease detection~\cite{CADsurvey}.
However, one of the major challenges in training deep learning models for medical purposes is the need for extensive, high-quality clinical annotation, which is time-consuming and costly.

Recently, CLIP~\cite{CLIP} and ALIGN~\cite{ALIGN} have shown the ability to perform vision tasks without any supervision. 
However, vision-language pre-training (VLP) in the CXR domain still lacks sufficient image-text datasets because many public datasets consist of image-label pairs with different class compositions.
MedCLIP~\cite{MedCLIP} attempted to a rule-based labler to use both image-text data and image-label data. 
However, it relies on the performance of the rule-based labeler and is not scalable to other diseases that the labeler cannot address.

In this paper, we propose a training method, \textit{CXR-CLIP}, that integrates image-text data with image-label data using class-specific prompts made by radiologists.
Our method does not depend on a rule-based labeler and can be applied to any image-label data.
Also, inspired by DeCLIP~\cite{DeCLIP}, we used Multi-View Supervision (MVS) utilizing multiple images and texts in a CXR study to make more image-text pairs for efficient learning.
In addition, we introduce two contrastive loss functions, named image contrastive loss (ICL) and text contrastive loss (TCL), to learn study-level characteristics of the CXR images and reports respectively.

The main contributions of this paper are summarized as follows. 1) We tackle the lack of data for VLP in CXR by generating image-text pairs from image-label datasets using prompt templates designed by radiologists and utilizing multiple images and texts in a study. 2) Two additional contrastive losses are introduced to learn discriminate features of image and text, improving image-text retrieval performances. 3) Performance of our model is validated on diverse datasets with zero-shot and few-shot settings.

\section{Related Work}
\textbf{Data Efficient VLP}
Recent studies~\cite{DeCLIP,SLIP} have proposed data-efficient VLP via joint learning with self-supervision. 
DeCLIP~\cite{DeCLIP} suggested MVS that utilizes image and text augmentation to leverage positive pairs along with other self-supervisions.
In CXR domain, GloRIA~\cite{GLORIA} aligned words in reports and sub-regions in an image for label efficiency, and BioVIL~\cite{BioVIL} combined self-supervision for label efficiency. 
We modify MVS as two distinct images and texts from a study and present self-supervised loss functions, ICL and TCL for efficient learning.

\textbf{Self-supervision within CXR study}
A CXR study could include several images in different views and two report sections: 'findings' and 'impression'. 
The impression section includes the differential diagnosis inferred from the findings section.
BioVIL~\cite{BioVIL} enhanced the text encoder by matching two sections during language pre-training.
MedAug~\cite{MedAug} shows that self-supervised learning by matching images in a study is better than differently augmented images.
We utilize both of multiple images and texts from a single study in VLP in an end-to-end fashion.

\textbf{Leveraging image-label data in VLP}
MedCLIP~\cite{MedCLIP} integrated unpaired images, texts, and labels using rule-based labeler~\cite{cheXpert}, which is less capable of retrieving the exact report for a given image due to the effect of decoupling image-text pairs.
UniCL~\cite{UniCL} suggested using prompts to leverage image-label dataset~\cite{deng2009imagenet}, considering the samples from the same label to be a positive pair.
To our knowledge, this is the first work to utilize prompting for training in CXR domain.

\begin{figure}[t!]
    \centering
    \includegraphics[width=\textwidth]{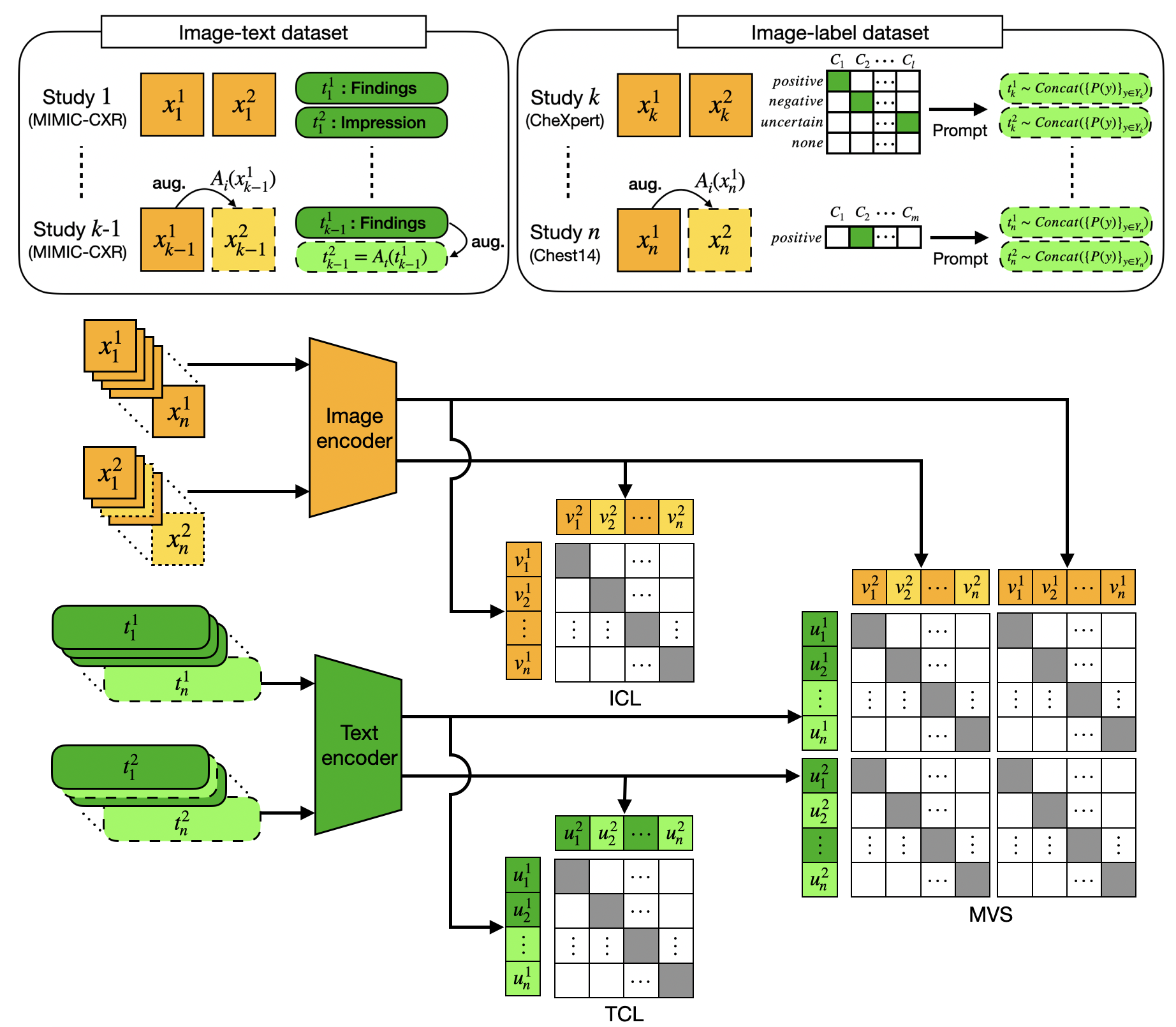}
    \caption{Overview of the proposed method with a training batch sampling $n$ studies, where each study has a pair of images ($x^1$, $x^2$) and a pair of text ($t^1$, $t^2$).
    If a study has one image or one text, data augmentation is conducted to make second examples.
    For the image-label data, two different prompts are generated from class labels as ($t^1$, $t^2$).
    Using sampled pairs, the encoders are trained with three kinds of contrastive losses (MVS, ICL, and TCL).}
    \label{fig:workflow}
\end{figure}

\section{Method}
CXR-CLIP samples image-text pairs from not only image-text data but also image-label data, and learns study-level characteristics with two images and two texts per study.
The overview of the proposed method is illustrated in Fig \ref{fig:workflow}.

\subsection{Data Sampling}\label{sampling}
We define a CXR study as $s = \{X, T\}$, where $X$ is a set of images, and $T$ is a set of "findings" and "impression" sections. 
The study of image-label dataset has a set of image labels $Y$ instead of $T$. 
For the image-label dataset, we make prompt-based texts $T = Concat(\{p \sim P(y)\}_{y\in Y})$, where $p$ is a sampled prompt sentence, $P(y)$ is a set of prompts given the class name and value $y$, and $Concat(\cdot)$ means concatenating texts.
The set of prompts is used to generate sentences such as actual clinical reports, taking into account class labels and their values (positive, negative, etc.), unlike the previous prompt~\cite{GLORIA} for evaluation which randomly combines a level of severity, location, and sub-type of disease.
Our prompts are available in Appendix.

We sample two images (${x^{1}, x^{2}}$) in $X$ if there are multiple images. Otherwise, we use augmented image $A_i({x}^{1})$ as $x^{2}$, where $A_i$ is image augmentation.
To leverage various information from different views in CXR (AP, PA, or lateral), we sample images from two distinct views as possible.
Similarly, we sample two texts (${t^{1}, t^{2}}$) in $T$ if there are both "findings" and "impression". Otherwise, we use augmented text $A_t(t^1)$ as $t^{2}$, where $A_t$ is text augmentation.
For the image-label data, we sample two prompt sentences as $t^{1}$ and $t^{2}$ from the constructed $T = Concat(\{p \sim P(y)\}_{y\in Y})$.

\subsection{Model Architecture}
We construct image encoder $E^i$ and text encoder $E^t$ to obtain global representations of image and text, and a projection layer $f^i$ and $f^t$ to match the size of final embedding vectors.

\textbf{Image Encoder} 
We have tested two different image encoders; ResNet-50~\cite{ResNet} and Swin-Tiny~\cite{SwinTrans} as follow~\cite{GLORIA,MedCLIP}.
We extract global visual features from the global average pooled output of the image encoder.
A linear layer is adopted to project the embeddings into the same size as text embeddings.
The normalized visual embedding $v$ is obtained by $v = f^i(E^i(x))~/~|| f^i(E^i(x)) ||$.
We denote a batch of the visual embeddings as $V = \{v\}_{i=1}^n$, where $n$ is a batch size.

\textbf{Text Encoder} 
We use BioClinicalBERT~\cite{ClinicalBert} model, which is the same architecture as BERT~\cite{Bert} but pre-trained with medical texts~\cite{MIMIC-3} as follow~\cite{GLORIA,MedCLIP}.
We use \textbf{[EOS]} token's final output as the global textual representation.
Also, a linear projection layer is adopted the same as the image encoder.
The normalized text embedding $u$ is denoted as $u = f^t(E^t(t))~/~|| f^t(E^t(t)) ||$.
We denote a batch of the text embedding as $U = \{u\}_{i=1}^n$ and ($v_i$, $u_i$) are paired.

\subsection{Loss Function}
In this section, we first describe CLIP loss~\cite{CLIP} and then describe our losses (MVS, ICL, TCL) in terms of CLIP loss.
The goal of CLIP loss is to pull image embedding and corresponding text embedding closer and to push unpaired image and text farther in the embedding space.
The InfoNCE loss is generally adopted as a type of contrastive loss, and CLIP uses the average of two InfoNCE losses; image-to-text and text-to-image. The formula for CLIP loss is given by
\begin{equation}
    L_{CLIP}(U, V) = -\frac{1}{2n} (\sum_{u_i \in U} \log \frac{\exp(v_i^T u_i/\tau)}{\sum_{v_j \in V}\exp(u_i^Tv_j/\tau)} + \sum_{v_i \in V} \log  \frac{\exp(u_i^T v_i/\tau)}{\sum_{u_j \in U}\exp(v_i^Tu_j/\tau)})
\end{equation}
, where $\tau$ is a learnable temperature to scale logits.

In DeCLIP~\cite{DeCLIP}, MVS uses four $L_{CLIP}$ loss with all possible pairs augmented views; ($x$, $t$), ($x$, $A_t(t)$), ($A_i(x)$, ${t}$) and ($A_i(x)$, $A_t(t)$).
We modify DeCLIP's MVS to fit the CXR domain by the composition of the second example.
DeCLIP only utilizes an augmented view of the original sample, but we sample a pair of the second image and text as described in \ref{sampling}.
We denote the first and the second sets of image embeddings as $U^1$, $U^2$, and text embeddings as $V^1$, $V^2$.
\begin{equation}
    L_{MVS} = \frac{1}{4} (L_{CLIP}(U^1, V^1) + L_{CLIP}(U^2, V^1) + L_{CLIP}(U^1, V^2) + L_{CLIP}(U^2, V^2))
\end{equation}

The goal of ICL and TCL is to learn modality-specific characteristics in terms of image and text respectively.
We design ICL and TCL as same as CLIP loss, but the input embeddings are different.
ICL only uses image embeddings; $L_{ICL} = L_{CLIP}(V^1, V^2)$ and TCL only uses text embeddings;
$L_{TCL} = L_{CLIP}(U^1, U^2)$.
ICL pulls image embeddings from the same study and pushes image embeddings from the different studies, so that, the image encoder can learn study-level diversity.
Similarly, TCL pulls embeddings of "findings" and "impression" in the same study or diverse expressions of prompts from the same label and pushes the other studies' text embeddings, so that the text encoder can match diverse clinical expressions on the same diagnosis.
Thereby, the final training objective consists of three contrastive losses balanced each component by $\lambda_{I}$ and $\lambda_{T}$, formulated by $L = L_{MVS} + \lambda_{I} L_{ICL} + \lambda_{T} L_{TCL}$.

\begin{table}[t]
\centering
\caption{The number of studies for each dataset and split in this paper}
\begin{tabular}{|c|ccc|cccc|} \hline
Data & \multicolumn{3}{c|}{Pre-training} &  \multicolumn{4}{c|}{Evaluation} \\
Split & \multicolumn{1}{c}{MIMIC-CXR}  & \multicolumn{1}{c}{CheXpert} & \multicolumn{1}{c|}{ChestX-ray14} & \multicolumn{1}{c}{VinDR} & \multicolumn{1}{c}{RSNA} & \multicolumn{1}{c}{SIIM} & \multicolumn{1}{c|}{Open-I} \\ \hline
Train & 222,628 & 216,478 & 89,696 & 12,000 & 18,678 & 8,422 & \-  \\
Valid & 1,808 & 233 & 22,423  & 3,000 & 4,003 & 1,808 & \- \\ \hline
Test & 3,264 & 1,000 & \- & 3,000 & 4,003 & 1,807 & 3,788 \\ \hline
\end{tabular}
\label{tab:dataset}
\end{table}

\section{Experiment}
\subsection{Datasets}
We used three pre-trained datasets and tested with various external datasets to test the generalizability of models. The statistics of the datasets used are summarized in Table~\ref{tab:dataset}.

\textbf{MIMIC-CXR}~\cite{MIMIC-CXR} consists of CXR studies, each with one or more images and free-form reports. 
We extracted "findings" and "impression" from the reports.
We used the training split for pre-training and the test split for image-to-text retrieval.

\textbf{CheXpert}~\cite{cheXpert} is an image-label data with 14 classes, obtained from the impression section by its rule-based labeler, and each class is labeled as positive, negative, uncertain, or none (not mentioned).
We used the training split for pre-training with class-specific prompts.
\textbf{CheXpert5x200} is a subset of CheXpert for 5-way classification, which has 200 exclusively positive images for each class. 
Note that only the reports of CheXpert5x200 are publicly available, but the reports of CheXpert are not.
Following the previous works~\cite{GLORIA,MedCLIP}, we excluded CheXpert5x200 from the training set and used it for test.

\textbf{ChestX-ray14}~\cite{ChestXray14} consists of frontal images with binary labels for 14 diseases. 
Prompts are generated by sampling 3 negative classes per study. We used 20\% of the original training set for validation, and the remaining 80\% for pre-training.

\textbf{RSNA pneumonia}~\cite{RSNA} is binary-labeled data as pneumonia or normal. We split train/valid/test set 70\%, 15\%, 15\% of the dataset following~\cite{GLORIA} for the external classification task.

\textbf{SIIM Pneumothorax}\footnote{https://siim.org/page/pneumothorax\_challenge} is also binary labeled as pneumothorax or normal. We split the train/valid/test set same ratio as RSNA pneumonia following~\cite{GLORIA} and used it for the classification task.

\textbf{VinDR-CXR}~\cite{VinDR-CXR} contains 22 local labels and 6 global labels of disease, which were obtained by experienced radiologists. We split the validation set from the original training set.
Of 28 classes, "other diseases" and "other lesions" classes were excluded. Then, only 18 classes having 10 or more samples within the test set were evaluated for the binary classification of each class as follow~\cite{Relaxing}.

\textbf{Open-I}~\cite{OpenI} is an image-text dataset. From each study, one of the report sections and one frontal-view image were sampled and used for image-to-text retrieval.

\subsection{Implementation Details}\label{detail}
We used augmentations $A_i$ and $A_t$ to fit medical images and reports.
For $A_i$, we resize and crop with scale [0.8, 1.1], randomly adapt CLAHE~\cite{CLAHE}, and random color jittering; brightness, hue ratios from [0.9, 1.1] and contrast, saturation [0.8, 1.2].
For $A_t$, to preserve clinical meaning, sentence swap and back-translation\footnote{https://huggingface.co/Helsinki-NLP} from Italian to English is used.
The image size and final-embedding size are set to 224 and 512 respectively as in previous work~\cite{MedCLIP}.
We set $\lambda_{I}$ and $\lambda_{T}$ to 1.0, 0.5 for balancing total loss.
Two encoders were trained for 15 epochs in a mixed-precision manner, early stopped by validation loss, and optimized by AdamW~\cite{ADAMW} with an initial learning rate 5e-5 and a weight decay 1e-4.
We used cosine-annealing learning-rate scheduler~\cite{CosWarmup} with warm-up for 1 epoch.
A training batch consists of 128 studies with 256 image-text pairs. 
We implemented all experiments on PyTorch with 4 NVIDIA V100 GPUs.

\begin{table}[t]
\centering
\caption{Comparison with state-of-the-art for zero-shot(ZS) or few-shot(10\%) classification tasks. M, C, and C14 mean MIMIC-CXR, CheXpert, and ChestX-ray14, respectively. C$^{*}$ means CheXpert with reports, which are not publicly available. ResNet50 (${R50}$) and SwinTiny (${SwinT}$) mean the image encoder used for each model.}
\label{tab:classification}
\begin{tabular}{|l|c|ccc|ccc|ccc|c|}
\hline
\multicolumn{1}{|c|}{} & Pre-train & \multicolumn{3}{c|}{VinDR-CXR} & \multicolumn{3}{c|}{RSNA} & \multicolumn{3}{c|}{SIIM} & C5x200 \\
\multicolumn{1}{|l|}{\multirow{-2}{*}{Model Name}} & Dataset & ZS & 10\% & 100\% & ZS & 10\% & 100\% & ZS & 10\% & 100\% & ZS-ACC \\ \hline
GloRIA$_{R50}$ & C* & 78.0 & 73.0 & 73.1 & 80.6 & 88.2 & 88.5 & 84.0 & 91.5 & 91.9 & 62.4$^*$ \\ \hline
CXR-CLIP$_{R50}$ & M & 78.8 & 82.1 & 82.2 & 83.3 & 88.5 & 89.2 & 85.2 & 88.3 & 90.5 & 56.2 \\
CXR-CLIP$_{SwinT}$ & M & 78.3 & 84.9 & 85.4 & 81.3 & 88.0 & 88.4 & 85.5 & 86.9 & 88.3 & 54.3 \\ \hline
MedCLIP$_{SwinT}$ & M,C & 82.4 & 84.9 & 85.1 & 81.9 & 88.9 & 89.0 & 89.0 & 90.4 & 90.8 & 59.2 \\
CXR-CLIP$_{R50}$ & M,C & \textbf{83.0} & 81.4 & 82.1 & 81.7 & 88.5 & 88.9 & 86.4 & 88.4 & 90.7 & 61.7 \\
CXR-CLIP$_{SwinT}$ & M,C & 82.7 & 86.1 & 86.7 & \textbf{84.5} & 88.1 & 88.8 & 87.9 & 89.6 & 91.2 & 60.1 \\ \hline
CXR-CLIP$_{R50}$ & M,C,C14 & 78.1 & 80.2 & 81.0 & 81.8 & 88.7 & 89.3 & 85.2 & 91.5 & 92.8 & 60.3 \\
CXR-CLIP$_{SwinT}$ & M,C,C14 & 78.9 & \textbf{88.0} & \textbf{89.0} & 80.1 & \textbf{89.2} & \textbf{89.8} & \textbf{91.4} & \textbf{92.9} & \textbf{94.0} & \textbf{62.8} \\ \hline
\end{tabular}
\end{table}

\begin{table}[t]
\centering
\caption{Comparison with state-of-the-arts for image-to-text retrieval. 
The notations of datasets and models are same to Table \ref{tab:classification}.
}
\begin{tabular}{l|c|ccc|ccc|ccc|c}
\multicolumn{1}{c|}{} & Pre-Train & \multicolumn{3}{c|}{CheXpert5x200} & \multicolumn{3}{c|}{MIMIC-CXR} &\multicolumn{3}{c|}{Open-I} & Total \\
\multicolumn{1}{c|}{\multirow{-2}{*}{Model Name}} & Dataset & R@1 & R@5 & R@10 & R@1 & R@5 & R@10 & R@1 & R@5 & R@10 & RSUM \\ \hline
GloRIA$_{R50}$ & C* & \textbf{17.8} & \textbf{38.8} & \textbf{49.9} & 7.2 & 20.6 & 30.3 & 1.5 & 4.4 & 6.5 & 177.0 \\ \hline
CXR-CLIP$_{R50}$ & M & 9.4 & 23.0 & 32.6 & 21.4 & 46.0 & 59.2 & \textbf{3.8} & 8.2 & \textbf{12.3} & \textbf{216.9} \\
CXR-CLIP$_{SwinT}$ & M & 8.4 & 21.5 & 30.2 & \textbf{21.6} & \textbf{48.9} & \textbf{60.2} & 3.6 & \textbf{8.3} & 11.5 & 214.2 \\ \hline
MedCLIP$_{SwinT}$ & M,C & 2.6 & 3.0 & 3.6 & 1.1 & 1.4 & 5.5 & 0.1 & 0.4 & 0.7 & 18.4 \\ 
CXR-CLIP$_{R50}$ & M,C & 5.5 & 19.2 & 27.4 & 20.2 & 45.9 & 58.2 & 3.5 & 8.2 & 12.0 & 200.1 \\
CXR-CLIP$_{SwinT}$ & M,C & 8.5 & 23.0 & 31.6 & 19.6 & 44.2 & 57.1 & 3.1 & \textbf{8.3} & 11.6 & 207.0 \\ \hline
CXR-CLIP$_{R50}$ & M,C,C14 & 5.7 & 18.0 & 28.3 & 19.7 & 44.4 & 56.4 & 2.3 & 6.7 & 10.1 & 191.6 \\
CXR-CLIP$_{SwinT}$ & M,C,C14 & 7.0 & 20.1 & 29.7 & 20.9 & 46.2 & 58.8 & 2.4 & 6.6 & 9.4 & 201.1
\end{tabular}
\label{tab:retrieval}
\end{table}

\subsection{Comparison with State-of-the-arts}
\textbf{Zero-shot and few-shot classification}
Table~\ref{tab:classification} shows performance on classification tasks of our models and state-of-the-art models.
To evaluate zero-shot classification fairly, we used evaluation prompts suggested from previous works~\cite{BioVIL,GLORIA,Relaxing}.
The evaluation prompts are available in Appendix.
We evaluate binary classification computed by Area Under ROC (AUC) and multi-class classification computed by accuracy (ACC).
Our ResNet model trained with MIMIC-CXR outperforms GloRIA~\cite{GLORIA} except for CheXpert5x200, as GloRIA trained with image-text pair in CheXpert.
Our SwinTiny model trained with MIMIC-CXR and CheXpert outperforms MedCLIP~\cite{MedCLIP}, which is the same architecture trained with the same datasets, in most of the metrics.
Adding more pre-training datasets by prompting image-label datasets tends to improve performance for classifications, while the SwinTiny CXR-CLIP pre-trained with three datasets, performs the best for most of the metrics.
More comparison with self-supervised models is available in Appendix.

\textbf{Image-to-text retrieval}
We evaluated image-to-text retrieval computed by $R@K$, the recall of the exact report in the top $K$ retrieved reports for a given image. (Table \ref{tab:retrieval})
While GloRIA~\cite{GLORIA} uses image-text pairs in CheXpert(C*) which is not available in public, CXR-CLIP uses image-text in MIMIC-CXR. So we adapt an external image-text dataset Open-I~\cite{OpenI} for a fair comparison.
GloRIA has the best performance on CheXpert but our model trained with MIMIC-CXR, which has similar amounts of studies to CheXpert, outperforms on Open-I.
MedCLIP almost lost the ability to retrieve image-text due to decoupling pairs of image and text during pre-training.
In CXR-CLIP, adding more image-label datasets such as CheXpert and ChestX-ray14 degrades the image-text retrieval performance, possibly because the contribution of the text in original reports was diluted.

\begin{table}[t!]
\caption{Ablations and comparison with CLIP~\cite{CLIP} and DeCLIP~\cite{DeCLIP}. Our augmentations effectively preserves clinical meaning than EDA. Our full methodology (CXR-CLIP) outperforms DeCLIP.}
\label{tab:ablation}
\setlength{\tabcolsep}{2pt}
\centering
\begin{tabular}{l|cccc|ccc|c}
\multicolumn{1}{c|}{\multirow{2}{*}{Method}} & \multicolumn{4}{c|}{CheXpert 5x200} & \multicolumn{3}{c|}{MIMIC-CXR} & Total \\
\multicolumn{1}{c|}{} & \multicolumn{1}{c|}{ACC} & R@1 & R@5 & R@10 & R@1 & R@5 & R@10 & RSUM \\ \hline
Vanila CLIP & \multicolumn{1}{c|}{58.9} & 4.4 & 14.4 & 22.6 & 17.3 & 41.2 & 52.6 & 152.5 \\ \hline
+ Study Level Sampling & \multicolumn{1}{c|}{58.7} & 4.6 & 15.1 & 23.2 & 17.8 & 42.5 & 54.2 & 157.4 \\
+ Augmentations & \multicolumn{1}{c|}{60.6} & 5.7 & 17.0 & 24.9 & 16.1 & 40.2 & 51.5 & 155.4 \\
+ MVS & \multicolumn{1}{c|}{61.2} & 5.4 & 17.1 & 24.7 & 16.3 & 40.6 & 53.3 & 157.4 \\
+ ICL & \multicolumn{1}{c|}{61.6} & \textbf{6.8} & \textbf{20.3} & 28.6 & 17.5 & 41.6 & 53.2 & 168.0 \\
+ TCL (CXR-CLIP) & \multicolumn{1}{c|}{\textbf{61.7}} & 6.2 & 18.2 & \textbf{29.1} & \textbf{19.6} & \textbf{44.8} & \textbf{56.6} & \textbf{174.5} \\ \hline
MVS of DeCLIP (EDA) & \multicolumn{1}{c|}{59.5} & 3.2 & 15.5 & 22.9 & 15.8 & 39.1 & 51.5 & 148.0 \\
MVS of DeCLIP (Our aug) & \multicolumn{1}{c|}{59.4} & 6.0 & 17.0 & 24.4 & 15.1 & 38.8 & 51.8 & 153.1 \\
DeCLIP (Our aug) & \multicolumn{1}{c|}{59.4} & 5.7 & 16.1 & 24.6 & 18.1 & 44.0 & 55.3 & 163.8
\end{tabular}
\end{table}

\subsection{Ablations}
For the ablation study, models with ResNet-50~\cite{ResNet} backbone were trained on MIMIC-CXR and CheXpert datasets and tested on zero-shot classification and image-to-text retrieval tasks with MIMIC-CXR and CheXpert5x200 datasets. 

We conducted two ablations shown in Table~\ref{tab:ablation}.
First, we analyzed the effect of each component of CXR-CLIP by adding the components to vanilla CLIP~\cite{CLIP} one by one.
To validate our data sampling closer, we divided the sampling method into three parts 1) study-level sampling 2) data augmentations 3) Multi-view and Multi-text sampling (MVS).
Our study-level sampling strategy improves performance compared to vanilla CLIP, which uses a naive sampling method bringing an image and corresponding report.
Additionally, the modified data augmentation to fit the CXR domain contributes to performance increment of classification, the similar performance on retrieval.
MVS slightly improves performances in both classification and image-text retrieval.
Adding more supervision (ICL and TCL) improves performance by utilizing better multi-views and multi-text inputs.
However, TCL drops the performance of recalls in CheXpert5x200, TCL could be hard to optimize variation of the radiologic report and prompt not diverse as images.

In the second ablation study, CXR-CLIP was compared to DeCLIP~\cite{DeCLIP} to confirm that our MVS using two image-text pairs per study is better than the MVS of DeCLIP which uses naively augmented images and texts.
We show that our text augmentation outperforms DeCLIP's text augmentation named EDA~\cite{EDA} in terms of image-to-text recall, which implies our text augmentation preserves clinical meaning.
The superiority of our MVS over DeCLIP's MVS confirms that using multiple images and texts from one study is better than using images and texts from augmented examples.
Also, our full methodology (CXR-CLIP) outperforms DeCLIP, suggesting that our method efficiently learns in the CXR domain more than DeCLIP.

\section{Conclusion}
We presented a framework enlarging training image-text pair by using image-label datasets as image-text pair with prompts and utilizing multiple images and report sections in a study.
Adding image-label datasets achieved performance gain in classification tasks including zero-shot and few-shot settings, on the other hand, lost the performance of retrieval tasks.
We also proposed loss functions ICL and TCL to enhance the discriminating power within each modality, which effectively increases image-text retrieval performance.
Our additional loss functions are designed to efficiently learn CXR domain knowledge along with image-text contrastive learning.

\bibliographystyle{splncs04}
\bibliography{paper2090}
\title{Supplementary Materials, CXR-CLIP: Toward Large Scale Chest X-ray Language-Image Pre-training}
\titlerunning{CXR-CLIP: Toward Large Scale Chest X-ray Language-Image Pre-training}
\author{Kihyun You\inst{1} \and
Jawook Gu\inst{1} \and
Jiyeon Ham\inst{1} \and
Beomhee Park\inst{1} \and
Jiho Kim\inst{1} \and
Eun K. Hong\inst{1} \and
Woonhyuk Baek\inst{1} \and
Byungseok Roh\inst{1}
}
% index{Kihyun, You}
% index{Jawook, Gu}
% index{Jiyeon, Ham}
% index{Beomhee, Park}
% index{Jiho, Kim}
% index{Eun, Hong}
% index{Woonhyuk, Baek}
% index{Byungseok, Roh}

\authorrunning{K. You et al.}
\institute{Kakaobrain, Seongnam, Republic of Korea  \\
\email{\{ukihyun, jawook.gu, jiyeon.ham, brook.park, tyler.md, amy.hong, wbaek, peter.roh\}@kakaobrain.com}\\}

\maketitle              % typeset the header of the contribution

\begin{table}[h!]
\caption{Comparison between our and GloRIA~\cite{GLORIA} prompt on cheXpert5x200~\cite{cheXpert}. Performance gain of the model not trained with prompt (MIMIC-CXR~\cite{MIMIC-CXR}) suggests that our prompts also worth to evaluate. Training with our prompts further improved performance.}
\centering
\begin{tabular}{|l|c|c|c|}
\hline
\multicolumn{1}{|c|}{} & Pre-train & GloRIA prompt & Our prompt \\
\multicolumn{1}{|l|}{\multirow{-2}{*}{Model Name}} & Dataset & C5x200 ACC & C5x200 ACC \\ \hline
CXR-CLIP$_{SwinT}$ & M &  54.3 & 56.1 \\ \hline
CXR-CLIP$_{SwinT}$ & M,C &  60.1 & 64.2\\ \hline
CXR-CLIP$_{SwinT}$ & M,C,C14 & \textbf{62.8} & \textbf{65.7} \\ \hline
\end{tabular}
\end{table}

\begin{table}[h]
\caption{Comparison with self-supervised models (REFERS~\cite{REFERS} and MRM~\cite{MRM}) in terms of classification tasks. We compared two-settings linear-probing and fine-tune whole visual backbone. All the models has ViT-base~\cite{ViT} backbone and are trained on MIMIC-CXR}
\centering
\begin{tabular}{|l|cc|cc|cc|}
\hline
\multirow{2}{*}{Model Name} & \multicolumn{2}{c|}{VinDR-CXR} & \multicolumn{2}{c|}{RSNA} & \multicolumn{2}{c|}{SIIM} \\
 & Linaer & Fine-tune & Linaer & Fine-tune & Linaer & Fine-tune \\ \hline
REFERS$_{ViT-B}$ & 83.6 & 90.1 & 86.7 & 87.9 & 81.3 & 89.5 \\
MRM$_{ViT-B}$ & 77.0 & 91.3 & 86.7 & 89.9 & 86.0 & \textbf{93.3} \\
CXR-CLIP$_{ViT-B}$ & \textbf{89.3} & \textbf{91.6} & \textbf{89.6} & \textbf{90.3} & \textbf{90.2} & 92.7 \\ \hline
\end{tabular}
\end{table}

\begin{table}[]
\caption{Evaluation prompts for zero-shot classification. For VinDR-CXR and SIIM, we use simple prompt in Jang et el.~\cite{Relaxing}, and we use prompt in BioVIL~\cite{BioVIL} for RSNA.}
\centering
\begin{tabular}{l|l|l}
Dataset  & Positive  & Negative   \\ \hline
VinDR-CXR, SIIM & \{classname\} & No \{classname\}  \\ \hline
RNSA-pneumonia & Findings suggesting pneumonia.        & No evidence of pneumonia.
\end{tabular}
\end{table}

\begin{table}[h!]
\caption{To compare BioVIL~\cite{BioVIL}, we train our ResNet models with image resolution 512, denoted CXR-CLIP$_{R50}^+$. RSUM is sum of recall$@k$, where $k=\{1,5,10\}$. Our models trained MIMIC outperforms BioVIL and CXR-CLIP$_{R50}^+$ generally outperforms CXR-CLIP$_{R50}$}
\centering
\begin{tabular}{|l|c|c|c|c|cc|c|c|}
\hline
\multirow{2}{*}{Model Name} & \multirow{2}{*}{\begin{tabular}[c]{@{}c@{}}Pre-train\\ Dataset\end{tabular}} & VinDR & RSNA & SIIM & \multicolumn{2}{c|}{CheXpert5x200} & MIMIC & OpenI \\ \cline{3-9} 
 &  & ZS(AUC) & ZS(AUC) & ZS(AUC) & \multicolumn{1}{c}{ZS(ACC)} & RSUM & RSUM & RSUM \\ \hline
BioVIL$_{R50}$ & M & - & 83.1 & - & - & - & - & - \\
CXR-CLIP$_{R50}$ & M & 78.8 & 83.3 & \textbf{85.2} & 54.0 & \textbf{65.0} & 126.6 & 25.3 \\
CXR-CLIP$_{R50}^+$ & M & \textbf{82.2} & \textbf{84.8} & \textbf{85.2} & \textbf{56.8} & 64.6 & \textbf{133.7} & \textbf{27.5} \\ \hline
CXR-CLIP$_{R50}$ & M,C & 83.0 & 85.0 & 86.4 & \textbf{61.7} & \textbf{52.1} & \textbf{124.3} & \textbf{23.7} \\
CXR-CLIP$_{R50}^+$ & M,C & \textbf{87.5} & \textbf{85.3} & \textbf{89.0} & 57.2 & 50.1 & 117.6 & 22.6 \\ \hline
CXR-CLIP$_{R50}$ & M,C,C14 & 78.1 & 81.8 & 85.2 & 60.3 & 52.0 & 120.5 & 19.1 \\
CXR-CLIP$_{R50}^+$ & M,C,C14 & \textbf{84.6} & \textbf{86.7} & \textbf{87.3} & \textbf{62.0} & \textbf{59.6} & \textbf{120.7} & \textbf{25.1} \\ \hline
\end{tabular}
\end{table}

\begin{table}[h!]
\caption{Default positive and negative templates for suggested prompts, as well as class-specific templates. $E$ is expressions for each class, $+$ means text concatenation, [$\cdot$] means random selection from the given list, and ( ) is blank text.}
\centering
\resizebox{\textwidth}{!}{
\begin{tabular}{|l|l|l|}
\hline
 & Positive templates & Negative templates \\ \hline
Default & \begin{tabular}[c]{@{}l@{}} {[}\{$E$\}., There is \{$E$\}.,\\ \{$E$\} is {[}present, seen, noted{]}.,\\ the presence of \{$E$\} is {[}seen, noted{]}. {]}\\ \end{tabular} & 
\multirow{2}{*}[1em]{\begin{tabular}[c]{@{}l@{}}{[}
{[}There is, ( ){]}  + {[}no \{$E$\}., \\ 
no radiographic evidence for \{$E$\}.,\\ 
no {[}visible, definite, obvious, appreciable, evident{]} \{$E$\}.,\\ 
no {[}convincing, definite, ( ){]} evidence of \{$E$\}., \\ 
no convincing signs of \{$E$\}.{]},\\ 
No \{$E$\} is {[}visible, present, noted{]}. {]}\end{tabular}} \\ %% Missing "{" and "}" here
 \cline{1-2}
% \end{tabular} \\ \hline
\begin{tabular}[c]{@{}l@{}}Edema \\ Pneumonia\end{tabular}& \begin{tabular}[c]{@{}l@{}} {[}Default Positive templates, \\ Findings are + {[}suggesting, \\ compatible with,  suggestive of, \\ representing{]} + \{$E$\}. {]} \end{tabular} & \\ \hline
%\begin{tabular}[c]{@{}l@{}}Default \end{tabular}\\ 
Cardiomegaly & \begin{tabular}[c]{@{}l@{}}{[}heart size, cardiac size, cardiac silhouette,\\  cardiac shadow, cardiac contour{]} \\ + {[}is, appears{]} + {[}enlarged, increased{]}.\\ \end{tabular} & \begin{tabular}[c]{@{}l@{}}{[}heart size, cardiac size, cardiac silhouette,\\  cardiac shadow, cardiac contour{]} \\ + {[}is, appears{]} \\ + {[}normal, within normal limits, unremarkable{]}.\\ \end{tabular} \\ \hline
\begin{tabular}[c]{@{}l@{}}Enlarged \\ Cardio-\\ mediastinum\end{tabular} & \begin{tabular}[c]{@{}l@{}}{[}{[}cardiomediastinal, mediastinal{]} silhouette,\\  {[}cardiomediastinum, mediastinum{]},\\ mediastinal contour{]}\\  + {[}is, appears{]} + {[}enlarged, widened{]}.\\ \end{tabular} & \begin{tabular}[c]{@{}l@{}}{[}{[}cardiomediastinal, mediastinal{]} silhouette,\\  {[}cardiomediastinum, mediastinum{]},\\ mediastinal contour{]}\\ + {[}is, appears{]} \\ + {[}normal, within normal limits, unremarkable{]}.\\ \end{tabular} \\ \hline
No Finding & \begin{tabular}[c]{@{}l@{}}{[}the lungs, both lungs, \\ the lung fields, both lung fields{]} + \\ {[}are clear, appear clear{]}.\\ \end{tabular} & \multicolumn{1}{c|}{-} \\ \hline
\end{tabular}
}
\end{table}

\begin{table}[h!]
\centering
\caption{Various expressions $E$ for classes using default templates. $+$, [$\cdot$] and ( ) are same as Table 4.}
\resizebox{\textwidth}{!}{
\begin{tabular}{|l|l|}
\hline
Class Name & Expressions $E$ \\ \hline
Atelectasis & [Atelectasis] \\ \hline
Consolidation & [Consolidation] \\ \hline
Edema & [Pulmonary edema] \\ \hline
Emphysema & [Emphysema, Emphysematous change] \\ \hline
Fibrosis & \begin{tabular}[c]{@{}l@{}} {[}{[}( ), pulmonary{]}+ {[}( ), fibrotic{]} + {[}scar, scarring{]}, \\ parenchymal + {[}scar, scarring{]}, fibrotic change{]}\end{tabular} \\ \hline
Fracture & {[}Fracture, Acute fracture{]} \\ \hline
Hernia & [Hernia, Herniation, Hiatal Hernia] \\ \hline
Infiltration & \begin{tabular}[c]{@{}l@{}} [{[}( ), pulmonary{]} + infiltration, infiltrate, \\
infiltrative + {[}density, opacity, process{]}] \end{tabular} \\ \hline
Lung Lesion & \begin{tabular}[c]{@{}l@{}} Pos: [lung lesion] \\ Neg: [{[}lung, pulmonary{]} + {[}nodule, mass, lesions, nodules or masses{]}] \end{tabular} \\ \hline
Lung Opacity & [pulmonary opacity] \\ \hline
Mass & [{[}pulmonary, lung{]} + mass] \\ \hline
Nodule & [{[}( ) , pulmonary{]} + {[}nodule, nodular opacity, nodular density{]}] \\ \hline
Pleural Effusion & [Pleural Effusion] \\ \hline
Pleural Other & [Pleural Abnormality] \\ \hline
Pleural Thickening & [Pleural Thickening, Thickened pleura] \\ \hline
Pneumonia & [Pneumonia] \\ \hline
Pneumothorax & [Pneumothorax] \\ \hline
Support Devices & [Support Devices] \\ \hline
\end{tabular}
}
\end{table}

\end{document}